# From Street View Imagery to Street Quality Indicators: Vision–Language Inference for the Suburban 15-minute City.

Joan Perez [a,*], Giovanni Fusco [b]

[a] *Urban Geo Analytics, France*

[b] *Université Côte-Azur-CNRS-AMU-Avignon Université, ESPACE, France*

* Corresponding author: jperez@urbangeoanalytics.com

## Abstract

Streetscape quality has become a central concern in contemporary urban planning, particularly within the framework of the pedestrian-friendly 15-minute city, where walkability and public-space quality are increasingly recognized as key determinants of urban performance. However, assessing streetscape qualities across large suburban and peri-urban territories remains challenging due to the time and resource demands of conventional field surveys. This paper presents a planning-oriented assessment of streetscape qualities in the north-eastern periphery of Nice (France) using the latest release of SAGAI (Streetscape Analysis with Generative AI), an open-source workflow that leverages vision–language models (VLMs) for large-scale streetscape analysis from Google Street View imagery. The new release addresses several limitations identified in the original framework through improved image acquisition, geographically consistent view generation, support for multiple VLM architectures, consensus-based inference, and an integrated analytical environment. The workflow is applied to several thousand street-level observations to evaluate qualities relevant to pedestrian-friendly urban environments: sidewalk presence, pedestrian entrance density and vegetation type. The resulting maps reveal that the desired streetscape qualities characterize only a fraction of today's suburban streetscapes, mainly in compact developments and traditional suburban faubourgs, while they are particularly lacking on residential hills. The analysis demonstrates the potential of contemporary VLMs to support urban diagnostics in extensive suburban territories where traditional fieldwork would be prohibitively time-consuming. Beyond the case study, the paper illustrates how recent advances in vision–language models can contribute to evidence-based planning by enabling scalable, flexible, and interpretable assessments of urban public-space quality.

## 1. Introduction

Streetscape assessment has become central to evidence-based urban planning, informing decisions about walkability, safety, commercial vitality, and public space quality (Gehl and Svarre, 2013; Harvey and Aultman-Hall, 2016; Ewing and Handy, 2009). More recently, the 15-minute city has been reframed as a planning strategy in which walkability and public space quality are considered even more critical than the mere presence of urban functions (Fusco et al. 2024). Consequently, streetscape assessment has become a key component of diagnostic analyses aimed at evaluating urban environments. Urban peripheries and suburbs present a particular challenge, as they often combine poor-quality streetscapes with large spatial extents that are difficult to assess through conventional fieldwork alone.

The emergence of street-level imagery platforms, particularly Google Street View (Anguelov et al., 2010), has opened new avenues for large-scale visual auditing of urban environments. However, translating these images into structured, interpretable indicators has traditionally required either labour-intensive manual annotation or task-specific computer vision models trained on extensive labelled datasets (Li & Ratti, 2018; Hosseini et al., 2022; Zhang et al., 2024).

Vision-language models (VLMs) have recently transformed this landscape by combining visual encoders with large language model decoders, enabling zero-shot interpretation of images through natural language prompts (Liu et al., 2024; Bai et al., 2025). The original SAGAI workflow (Streetscape Analysis with Generative AI, Perez & Fusco, 2025), released as open-source code, demonstrated that VLMs can be deployed for streetscape scoring using open-access geospatial data and lightweight computing environments. The framework can therefore be applied in any region where *Google Street View* imagery is available, without requiring task-specific training or additional proprietary datasets. By formulating analytical tasks directly through prompts, the system enables the automated extraction and mapping of streetscape indicators on the fly, allowing users to generate spatial datasets for diverse urban analysis tasks without modifying the underlying model. Two exploratory case studies in the outskirts of Nice (France) and Vienna (Austria) showed mid to strong performances for binary urban–rural classification (92% accuracy), storefront detection (64%), and sidewalk width (54%).

However, SAGAI v1.0 also revealed significant limitations. It relied on a single model checkpoint (LLaVA v1.6 Mistral-7B), an adapter-based vision–language architecture that connects a pretrained vision encoder to a separate language model rather than a natively trained multimodal model. As a result, the framework provided no mechanism to compare performance across alternative VLM architectures. This limitation constrained the study to the capabilities of a single architecture and prevented the authors from employing assessment tasks targeting relevant urban qualities that require intensive reasoning. Moreover, the image acquisition module used fixed cardinal headings (0°, 90°, 180°, 270°), which were not aligned with street geometry, making "left" and "right" views meaningless for directional analysis. Points were placed at theoretical positions without verification against actual Street View camera locations, leading to spatial mismatches and duplicate downloads where parallel sidewalks and roads were digitized separately in *OpenStreetMap*. The inference module lacked consensus mechanisms to address VLM output stochasticity, and four separate notebooks required manual file management between modules.

This paper presents a planning-relevant analysis of the urban qualities of streetscapes in the north-eastern periphery of Nice (France) to assess their fine-grained characteristics within the context of the pedestrian-friendly 15-minute city. The analysis was made possible by the latest iteration of SAGAI, which overcomes each of the above-mentioned limitations.

The principal methodological developments are: (1) Integration of UVLM (Perez and Fusco, 2026), imported as a Python package[1], as the inference engine, providing access to 24 VLM checkpoints with consensus validation and chain-of-thought reasoning; (2) A canonical bearing algorithm that normalizes street orientation to [0°, 180°), producing geographically consistent directional views; (3) A metadata-driven image acquisition pipeline that eliminates duplicates, filters indoor photographs, and snaps points to verified camera locations; (4) A consolidated single-notebook architecture with interactive polygon-based study area definition; and (5) Per-view aggregation enabling analysis of streetscape asymmetries.

The remainder of this paper is organized as follows. Section 2 reviews the theoretical foundations of vision-language models as inferential systems in urban studies. Section 3 presents the methodology behind the last iteration of SAGAI. Section 4 describes the case study and the investigated streetscape qualities. Section 5 reports results. Section 6 discusses implications while section 7 concludes the paper.

## 2. Theoretical foundations: from detection to inference

### 2.1. The limits of discriminative computer vision

Traditional computer vision approaches to streetscape analysis operate within a discriminative paradigm: they classify, segment, or detect objects that are visually present and explicitly labeled in training data (Girshick et al., 2014; Redmon et al., 2016). Semantic segmentation models assign class labels to every pixel, producing precise delineations of buildings, vegetation, roads, and sky. Object detectors identify and localize discrete entities within bounding boxes. These approaches have powered influential streetscape studies, from Treepedia's Green View Index (Li & Ratti, 2018) to Urban Visual Intelligence's hierarchical analysis (Zhang et al., 2024).

[1] https://github.com/perezjoan/UVLM

However, discriminative models face three fundamental constraints. First, they are bound by their training ontology: they can only recognize categories in their training data. Second, they produce categorical or geometric outputs but cannot generate nuanced semantic assessments requiring compositional reasoning. Third, and most critically for urban analysis, they can only report what is explicitly visible. A segmentation model can identify that 30% of the image is "building," but it cannot infer that a building entrance is behind a parked truck, that a sidewalk continues beyond a construction barrier, or that commercial signage density suggests a high-street environment even when individual storefronts are partially occluded.

## 2.2. Vision-language models as inferential systems

Vision-language models represent a paradigmatic shift from detection to inference. By coupling a visual encoder, typically a CLIP-based Vision Transformer (Radford et al., 2021), with a large language model decoder, VLMs can reason about visual content using the world knowledge embedded in their language model component. This architecture enables what we term inferential visual analysis: the ability to go beyond pixel-level detection to generate contextual interpretations, causal hypotheses, and structured assessments from images.

A key capability that distinguishes VLMs from discriminative models is amodal reasoning: the ability to infer the presence or properties of objects that are partially or fully occluded. Gestalt psychology has long established that human visual perception completes partially hidden objects through amodal completion (Köhler, 1929; Kellman and Spelke, 1983). Recent benchmarks show that multimodal large language models can perform analogous operations: CAPTURe (Pothiraj et al., 2025) tests VLMs on counting objects behind occluders, while O-Bench (Liu et al., 2025) evaluates occlusion perception across five distinct task types. Both reveal that VLMs can leverage contextual patterns to infer unseen spatial relationships, though they still lag behind human performance.

For streetscape analysis, this inferential capacity has direct practical implications. When asked "How wide is the sidewalk?" in an image where a parked car partially obscures the pavement, a VLM can draw on contextual cues; the visible portion, the car's proportions as a reference object, adjacent road markings, to produce a reasonable estimate, whereas a segmentation model would simply leave occluded regions unresolved. Similarly, when asked "Are there commercial storefronts?" where a delivery van blocks the ground floor, the model can infer commercial activity from signage above the van, adjacent building types, or outdoor seating. This capacity for contextual inference makes VLMs suitable for holistic streetscape assessment that traditionally required human experts walking the street (e.g. Clarke et al., 2010; Rundle et al., 2011).

Recent validation studies support this. Mushkani (2025) shows that VLMs demonstrate reasonable alignment with human judgments on objective street attributes but struggle with subjective appraisals, while Yao (2026) demonstrate that VLMs can simultaneously assess multiple urban perception dimensions. These findings converge on a key insight: VLMs are most effective for structured, objective scoring tasks, precisely the design target of SAGAI.

## 2.3. A rapidly expanding field

In parallel with these validation efforts, the application of VLMs in geographic and urban research has expanded rapidly. Since the publication of the original SAGAI workflow (Perez & Fusco, 2025), the application of VLMs in geographic research has accelerated. Lan (2025) developed an interpretable multimodal framework for perceptual urban diagnostics, using VLMs to capture qualitative dimensions of walkability and safety beyond what GIS-based indicators can express. Kim et al. (2025) built *StreetLens*, a system of AI agents that replicate protocol-driven visual audits from Street View imagery at scale. Esposti et al. (2025) applied vision-based quality metrics across six Italian metropolitan areas, illustrating how such workflows can support city-wide comparative analysis.

The scope of VLM applications in urban research has also broadened. Zhang et al. (2026) proposed the Urban Landscape Emotion Analysis Framework (ULEAF), which combines convolutional neural networks with multimodal semantic matching models (CLIP and DeepSentiBank) to extract emotional and semantic information from social media images of urban parks. Their framework links visual landscape features with affective perception, enabling large-scale analysis of how environmental characteristics relate to positive or negative emotional responses. Palomares Avena (2025) coupled graph machine learning with a multi-agent framework that integrates LLMs to assess urban walkability, while Li et al. (2025) developed *BuildingView* to perform building annotations on building exteriors using LLMs.

Collectively, these works mark a transition from proof-of-concept to operational deployment of VLMs in urban analysis. The current version of SAGAI responds to this momentum by addressing practical challenges identified in this first wave: consolidating the workflow into a single reproducible notebook, supporting multi-model inference through the UVLM package (Perez and Fusco, 2026), and introducing features such as canonical bearing computation, metadata-based point snapping, panorama deduplication, and view-specific aggregation that make large-scale streetscape assessment more robust and analytically flexible.

# 3. Methodology: the SAGAI workflow

SAGAI is implemented as a single notebook containing six sequential code blocks, designed for Google Colab. The six blocks correspond to: (1) interactive study area definition and point generation; (2) image acquisition with metadata-driven quality control; (3–5) UVLM-based inference; and (6) spatial aggregation with cartographic output.

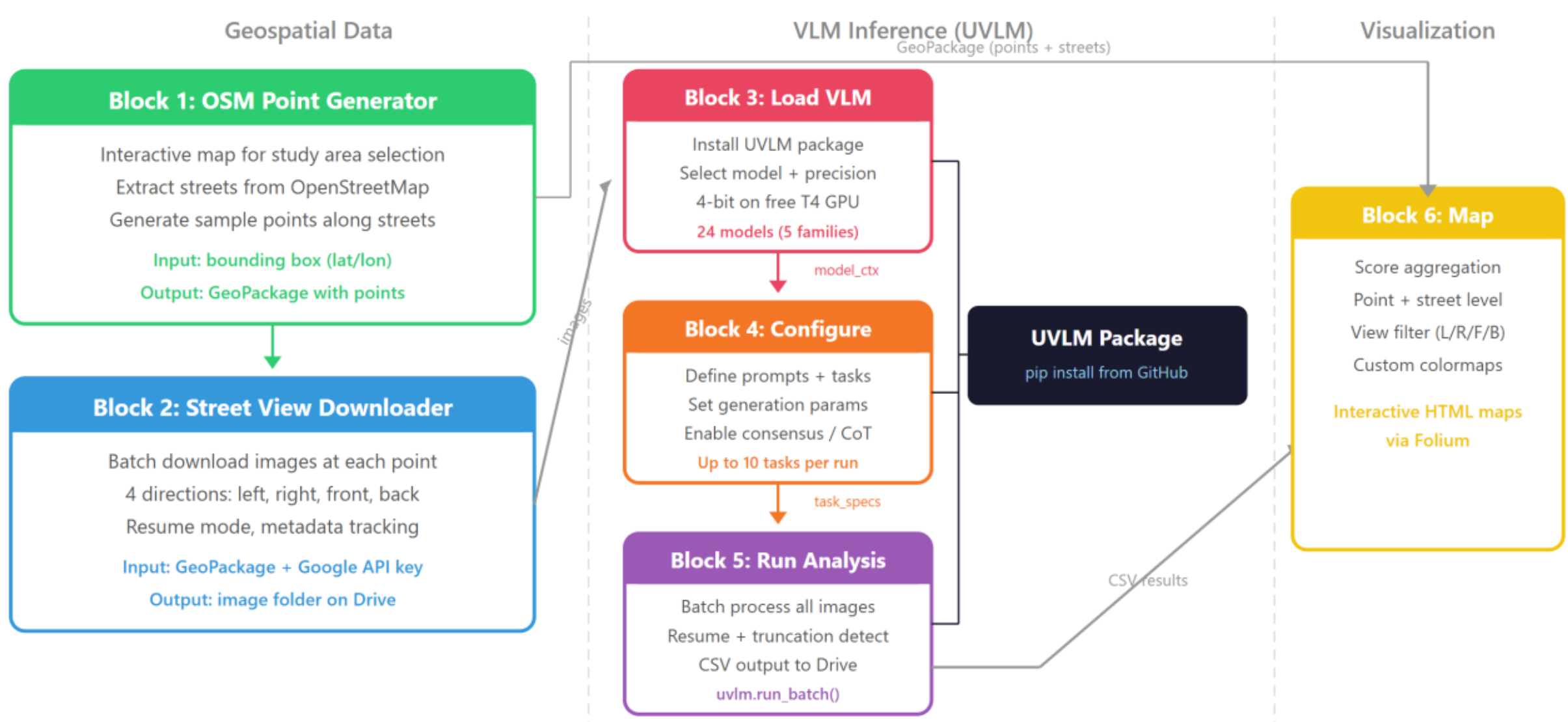

***Fig. 1.*** *The SAGAI architecture. A single notebook with six sequential blocks.*

### 3.1. Block 1: Interactive study area definition and point generation

Block 1 replaces the original bounding-box input with an interactive ipyleaflet map allowing users to draw arbitrary polygons. This enables analysis of irregularly shaped neighborhoods or administrative boundaries. It also allows the selection of very small study areas such as individual streets, short street segments, or predefined routes (e.g., parcours)**,** making the framework adaptable to corridor-based analyses or targeted site assessments. The module downloads the street network via OSMnx (Boeing, 2017). A configurable highway type filter allows excluding road categories (*footway, cycleway, path, steps*) that may generate parallel or redundant geometries. The projected CRS is automatically determined through UTM zone estimation. Points are distributed along the network at user-defined spacing (default 30 m) with configurable intersection offsets (default 15 m). Street and name attributes are preserved in the output GeoPackage.

### 3.2. Block 2: Metadata-driven image acquisition

Block 2 implements a three-phase pipeline addressing the original spatial imprecision, duplicate downloads, and indoor photo contamination. Before any API interaction, bearings are computed from original on-street positions—critical because post-snap positions may shift 20–30 m off-street. Each street segment is subdivided into sub-segments of configurable maximum length (default 50 m). For each point, the nearest sub-segment is identified, and the node-to-node bearing is computed. The raw bearing [0°, 360°) is normalized to the canonical range [0°, 180°): if $b \geq 180°$, subtract 180°. This ensures a street digitized A-to-B (≈0°) and B-to-A (≈180°) produce identical canonical bearings, so “left” (bearing − 90°) and “right” (bearing + 90°) always refer to the same geographic side.

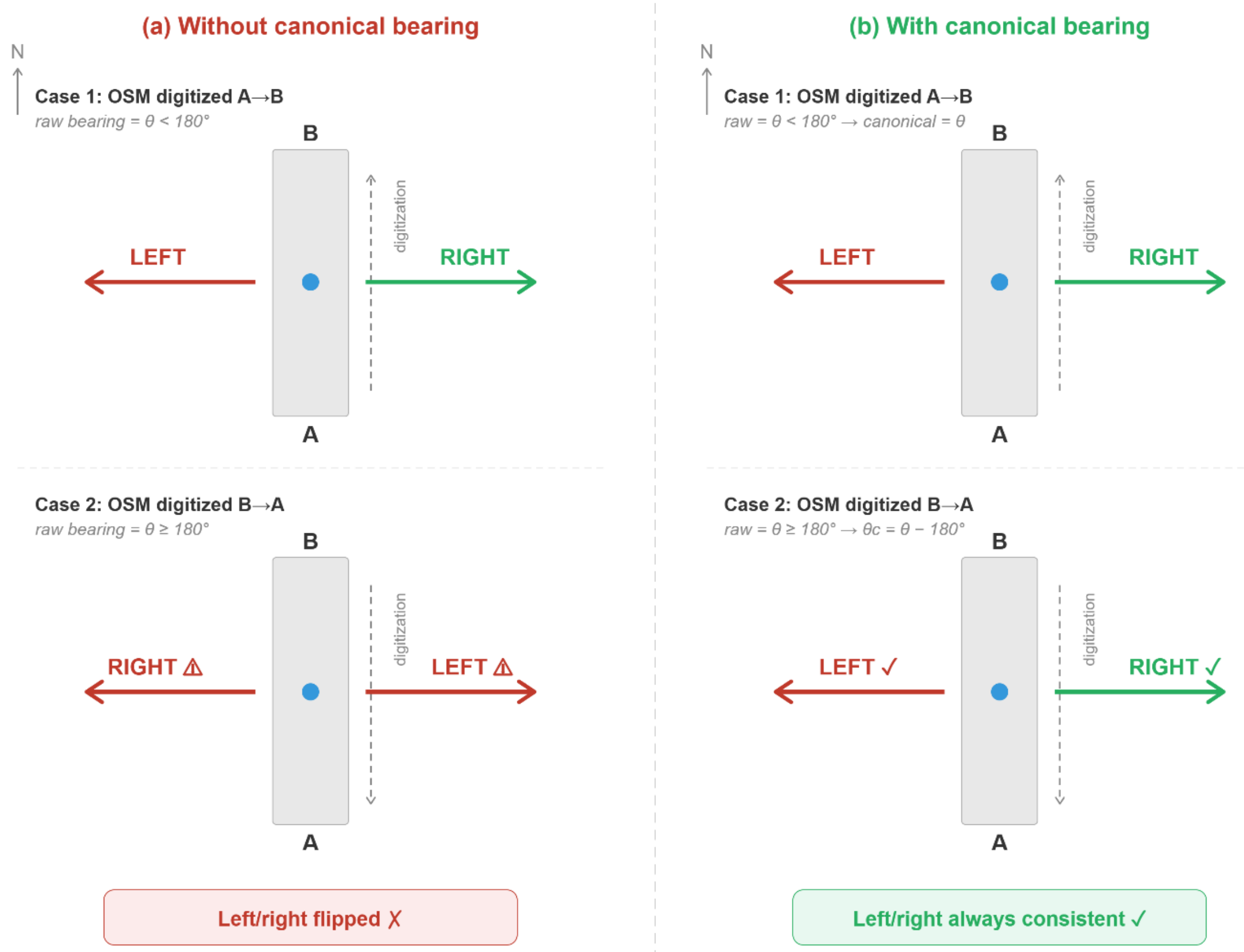

***Fig. 2.*** *Canonical bearing normalization. (a) Without normalization, left/right assignments depend on OSM digitization direction, flipping sides for the same physical street. (b) Normalizing to [0°, 180°) ensures left and right always refer to the same geographic side: if $\theta \geq 180°$, then $\theta c = \theta - 180°$.*

For each point, the free Google Street View Metadata API is queried, returning availability status, exact camera lat/lon, a unique pano_id, and a copyright field. Four filters are applied sequentially: (1) points with no imagery are removed; (2) images whose copyright does not contain “Google” are excluded, eliminating indoor 360° contributor photographs; (3) points where the distance to the camera exceeds a threshold (default 30 m) are removed; (4) points sharing the same *pano_id* are deduplicated. Surviving points are snapped to verified camera positions.

Images are downloaded only for surviving points, using the *pano_id* directly in API requests. Four views are computed from the canonical bearing: front, back, left, right. Users may select subsets. A real-time progress bar displays point count, image statistics, and ETA.

### 3.3. Blocks 3 to 5: UVLM-based vision-language inference

The inference engine is UVLM (Perez and Fusco, 2026), imported as a Python package and installed automatically from its GitHub repository at notebook startup. This represents a significant improvement, allowing analysts to choose from different VLMs with varying computational requirements according to the tasks to be performed. UVLM provides a unified interface to 24 checkpoints (1B–110B) across five model families (LLaVA-NeXT, Qwen2.5-VL, Qwen3-VL, InternVL3.5, and Gemma 4), abstracting away all architecture-specific loading and prompting logic. Key features integrated into SAGAI through this dependency: (1) a multi-task prompt builder supporting numeric, category, boolean, and text responses; (2) consensus validation via majority voting across repeated inferences; (3) advanced reasoning mode with chain-of-thought prompting and structured answer extraction; (4) truncation detection monitoring generated token counts in real time. See Perez and Fusco (2026) for implementation details.

### 3.4. Block 6: Spatial aggregation and cartographic output

Block 6 aggregates image-level scores to point and street levels. The image mode (4 views, left+right, or front+back) is auto-detected from CSV filename suffixes. A view filter restricts which directions enter the aggregation; seven options ranging from single sides (left only, right only) to combinations (left+right, front+back) or all views. This filter operates before statistical computation, so selecting "left only" produces scores reflecting exclusively one geographic side of each street, meaningful only because canonical bearings guarantee consistent side assignment.
For each point, four statistics are computed: mean, sum, median, and variance. Street-level scores are obtained by grouping points by segment. The user selects the display statistic (mean, sum, or median) and one of 11 color ramps, applied to both static matplotlib maps (points and streets side by side) and optional interactive folium HTML maps. The points map embeds Street View thumbnails in clickable popups alongside individual image scores; the streets map shows segment name, street type, and aggregated score. Both are saved as self-contained HTML files to Google Drive. All spatial outputs are also written to a GeoPackage with point and street layers for use in GIS software.

## 4. Case study

### 4.1. Study area

The study area is investigated within the emc2 project (Fusco et al., 2024), which aims to assess the potential of peripheral and suburban environments to support a pedestrian-friendly 15-minute city. Fusco and Picard (2025) previously conducted a detailed field survey of the faubourg of Drap, covering an area of approximately 1 km². The present study explores the possibility of scaling this analysis to a much larger territory located to the north-east of Nice (France), encompassing approximately 10 km² and a population of 28,000 inhabitants.
The study area includes a section of the narrow Paillon River valley and its surrounding hillsides, comprising the housing estate of L'Ariane (about 11,000 inhabitants, within the municipality of Nice), the town of La Trinité (about 11,000 inhabitants), the town of Drap—including both its historic faubourg and the northern development of La Condamine (about 5,500 inhabitants)—and the historic village of Cantaron (about 500 inhabitants).
Overall, the area constitutes a transition zone characterized by considerable morphological diversity. It includes dense urban fabrics along major arterial roads, residential neighbourhoods of varying densities, peri-urban developments composed of detached houses and garden plots, and mixed-use segments of the Paillon valley floor. This diversity makes the area a suitable testbed for evaluating streetscape indicators across a wide range of urban conditions, while its spatial extent provides a compelling case for the use of SAGAI as a scalable alternative to conventional field-based assessment.

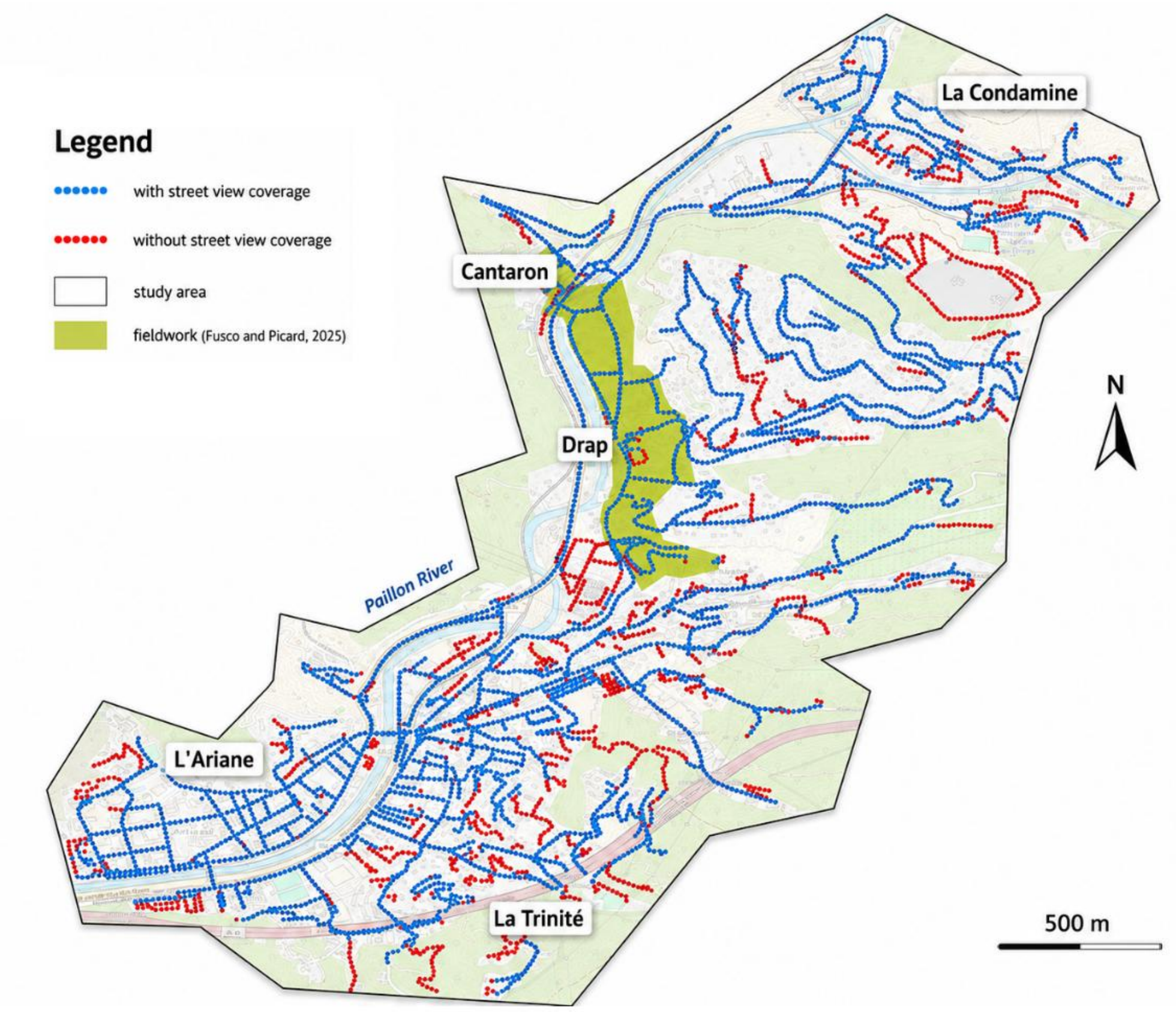

***Fig. 3.*** *Study area and distribution of sampling points with and without Street View coverage.*

## 4.2. Sampling and image acquisition

Block 1 generated 5,651 sampling points distributed at 20 m intervals with an offset of 5 m along the street network extracted from OpenStreetMap. After metadata-driven filtering in Block 2 (panorama snapping, duplicate removal, indoor photo exclusion, and distance thresholding), 4,036 points retained valid Street View coverage, representing 71.4% of the original sample (**fig.3**). The remaining 1,615 points (28.6%) had no available Google Street View imagery, typically corresponding to recently built streets, dead-end residential lanes, or pedestrian paths where the Street View vehicle has not operated. Only left and right views were downloaded, producing 8,072 images in total. This two-view configuration was chosen to capture the street frontage on both sides, which is the relevant observation unit for all four indicators used in this study. Front and back views, which capture the street axis rather than the frontage, were excluded.

## 4.3. Scoring tasks and model configuration

Four streetscape features were defined using the UVLM prompt builder, each targeting a distinct dimension of street-level morphology:

**Sidewalk presence** (boolean). The model detects whether a sidewalk or protected pedestrian path borders the visible street frontage. Both raised pavements and at-grade protected paths are considered. A sidewalk that does not span the entire frontage is still counted as present.

**Pedestrian entrance count** (numeric). The model counts all pedestrian entrances to the plots abutting the visible street frontage. This includes building doors and pedestrian gates in fences or walls, but excludes garage doors and vehicle-only gates unless they are the sole access point and bear a doorbell, street number, or mailbox. Entrances partially hidden by foreground objects are counted if sufficient visual cues are present.

**Vegetation score** (numeric, 1 to 6). The model classifies the vegetation regime along the street frontage using a six-level ordinal scale combining the degree of vegetation on abutting plots and the presence of canopy-forming street trees. The levels are ranked from most to least vegetated as follows: (1) highly vegetated abutting plots with canopy-forming street trees; (2) highly vegetated abutting plots without canopy-forming street trees; (3) partially vegetated abutting plots with canopy-forming street trees; (4) partially vegetated abutting plots without canopy-forming street trees; (5) little or no vegetation on abutting plots but with canopy-forming street trees; and (6) little or no vegetation either on abutting plots or along the street. Winter leaflessness is accounted for in the prompt: deciduous trees and shrubs are classified based on their trunk and branch structure.

**Street frontage length** (numeric, meters). The model estimates the total linear length of the street-facing boundary visible in the image, from the left edge to the right edge of the visible frontage, measured up to the first clearly visible perpendicular intersection. The prompt instructs the model to select a standard-size reference object (cars, parking bays, doors, sidewalk slabs, or building stories) and use it consistently to calibrate the estimate, accounting for perspective foreshortening.

All four prompts share a common preamble defining the concept of street frontage and its extent. The use of the UVLM package was essential for identifying the most suitable VLMs in terms of computational cost and accuracy. Accordingly, the first three tasks (sidewalk detection, pedestrian entrance counting, and vegetation classification) were performed using Qwen2.5-VL-32B-Instruct in standard mode. Using this model, Perez & Fusco (2026) reported a proximity score[2] of 80.8% for sidewalk detection, 92.3% for pedestrian entrance counting, and 89.8% for vegetation classification. Street frontage length estimation was performed using Qwen2.5-VL-7B-Instruct in reasoning mode, as this task requires spatial calibration against reference objects. Perez & Fusco (2026) reported a proximity score of 70.7% for this task. All runs used an A100 GPU. A single 32B standard run over the full image set takes approximately 2.5 hours; a 7B reasoning run takes approximately 35 hours, reflecting the much larger token budget required for chain-of-thought generation. No consensus validation was used; each image was scored once per task. The full prompts related to the four indicators analysed in this paper are detailed in **Appendix 1**.

## 4.4. Street-quality indicators

SAGAI outputs and GIS post-processing were used to derive three streetscape indicators for each junction-to-junction street segment.

**Street vegetation.** Vegetation scores obtained from the ordinal frontage typology were treated as numerical values and averaged across all left- and right-facing images associated with the segment.

**Street sidewalk coverage.** Sidewalk presence was modelled as a Bernoulli variable observed on each streetscape image. The proportion of images showing a sidewalk therefore provides an estimate of the probability of encountering a sidewalk at any point along the segment.

**Street constitutedness.** Constitutedness (Hillier 1996) was measured as entrance density, expressed as the number of pedestrian-accessible entrances per 100 m of street frontage. Following Hillier (1996) and Gehl (2011), entrance density is used as a proxy for the degree of interaction between buildings and public space. Operationally, the indicator is computed as the ratio between the total number of entrances identified on all street-front images and the cumulative length of the corresponding frontages.

[2] Proximity score measures prediction accuracy on a continuous scale from 0 to 1, awarding partial credit for near-miss predictions rather than requiring exact matches; for each prediction, the score decreases linearly with the absolute error $|y_i - \hat{y}_i|$, normalized by a range parameter R, reaching zero once the error equals or exceeds R.

**Quality streets.** To evaluate streetscape suitability for a pedestrian-friendly suburban 15-minute city, the three indicators were combined using threshold values. High constitutedness reflects active interfaces between buildings and public space. While Gehl (2010) suggests a threshold of six entrances per 100 m for moderately active urban frontages, a lower threshold of four entrances per 100 m was adopted to account for the lower densities typical of suburban environments. Sidewalk coverage was considered satisfactory when at least 75% of observations indicated the presence of a sidewalk, acknowledging that suburban streets frequently provide pedestrian infrastructure on only one side of the carriageway.
Vegetation was assessed differently. Rather than promoting high vegetation levels in all contexts (including dense commercial streets where this may be impractical) the objective was to exclude environments almost entirely devoid of vegetation. A segment was therefore considered satisfactory when its average vegetation score was equal to or lower than 5, thereby excluding situations characterized by negligible vegetation on adjacent plots and the absence of canopy-forming street trees.
Street segments satisfying all three criteria were classified as possessing sufficient streetscape quality to support pedestrian-friendly environments within the study area.

## 5. Results

The final spatial dataset for the study area covers 4,036 scored points distributed across 1,505 street segments for a total of 110.9 km of street length. Results are mapped in Fig. 4.

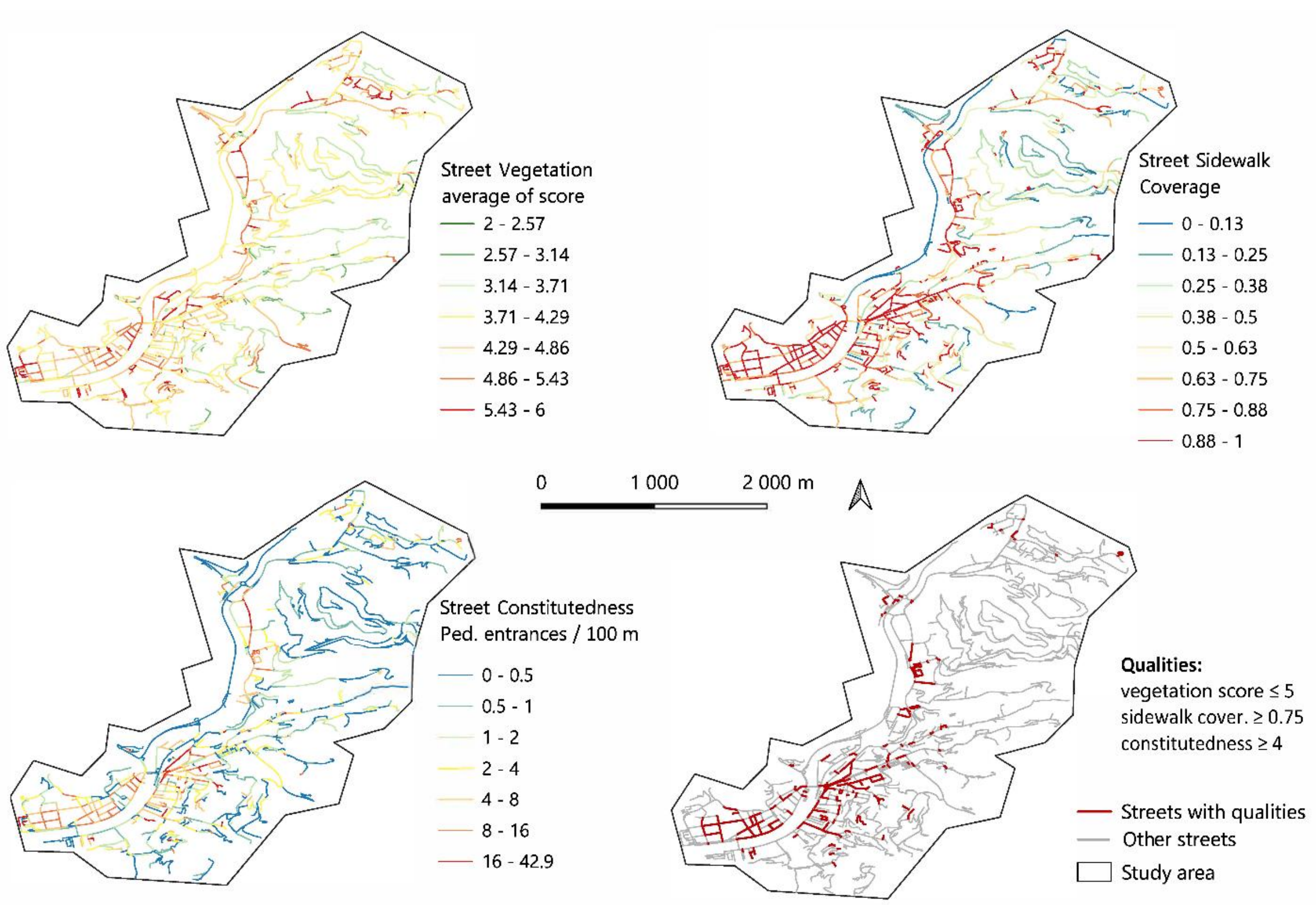

***Fig. 4.*** *Street-quality indicators in the north-eastern periphery of Nice (France).*

Street vegetation (upper left map) is generally abundant across the study area, especially in the residential neighbourhoods located on the eastern hillsides of the Paillon valley, where average vegetation scores are often below 3. More densely built environments, notably in La Trinité, still retain some vegetation along their streets. In contrast, several street segments within the L'Ariane housing estates exhibit almost no vegetation. Similar conditions are observed along the main axis of Drap's faubourg and in the commercial district of La Condamine.
The spatial distribution of sidewalk coverage (upper right map) is almost the inverse of vegetation. Coverage is highest in L'Ariane and remains relatively good in La Trinité, Drap's faubourg and the historic village of Cantaron. By contrast, La Condamine, particularly its commercial district, combines limited vegetation with poor sidewalk provision, reflecting a strongly utilitarian and car-oriented urban layout.
Street constitutedness (lower left map) reaches its highest values along the main streets of La Trinité and the northern section of Drap's faubourg, exceeding 16 pedestrian-accessible entrances per 100 m. The qualitative contrast between the northern and southern sections of Drap's main street had already been documented through field surveys by Fusco and Picard (2025). Constitutedness remains acceptable in most parts of L'Ariane and central La Trinité, but declines markedly in the residential hillsides and in both the residential and commercial sectors of La Condamine.
Combining the three indicators makes it possible to identify street segments meeting the adopted suitability criteria (lower right map). These represent only 10% of the street network (11.0 km) and are concentrated in the central areas of L'Ariane and La Trinité. Good streetscape quality is also found in the historic centres of Drap and Cantaron. By contrast, Drap's main axis is penalized by limited vegetation, with only a short segment in its northern section (close to the historic core) meeting all criteria. Streets in La Condamine and on the surrounding hillsides generally perform poorly due to inadequate sidewalk provision, low constitutedness, and, in the case of the commercial district, a near absence of vegetation.

## 6. Discussion

Although the analysis considers only three dimensions of pedestrian street quality and adopts thresholds adapted to suburban contexts, it nevertheless highlights the considerable challenge of creating a pedestrian-friendly 15-minute city in this section of the Paillon valley, on the north-eastern outskirts of Nice. Strategies focused exclusively on increasing functional mix and reducing travel distances are unlikely to succeed unless the quality of public space is simultaneously improved, encouraging walking and supporting social interaction. At the same time, the analysis reveals several clusters of high-quality streets within the most densely populated sectors, notably in L'Ariane, La Trinité, Drap and Cantaron. By identifying the specific factors responsible for poor performance, the results also provide a basis for targeted urban design interventions and site-repair strategies in the sense proposed by Alexander et al. (1977). Examples include increasing street-tree cover along Drap's main axis and in parts of L'Ariane, extending sidewalk provision on the residential hillsides, and rethinking the commercial district of La Condamine according to a more pedestrian-oriented paradigm.
The deficiencies identified by the analysis are particularly critical when they affect main streets, which constitute the social and functional backbone of the suburban 15-minute city (Fusco et al., 2024). Future applications should therefore distinguish more explicitly between main streets and ordinary streets. One promising avenue would be to combine the present assessment with approaches such as those proposed by Araldi et al. (2026), which identify streets with high configurational pedestrian centrality in suburban environments. Such an integration would make it possible to evaluate not only the quality of individual streetscapes but also their strategic importance within the wider urban structure.
More broadly, the SAGAI framework could be extended to assess additional dimensions of public-space quality. Potential indicators include the characteristics of building façades and their interfaces with public space, as proposed by Fusco and Picard (2025), as well as the presence of pedestrians and vehicles within streetscapes. The latter would require particular methodological attention, since observed levels of activity depend on the time of image acquisition and would therefore necessitate a more temporally representative sample of street-level imagery. Together, these developments would contribute to a richer and more comprehensive evaluation of the conditions supporting walkability and everyday urban life in suburban contexts.

## 7. Conclusion

Beyond the substantive findings for the Paillon valley, this study demonstrates the capacity of the latest SAGAI workflow to assess planning-relevant streetscape qualities at a scale that would be difficult to achieve through conventional fieldwork alone. The analysis covered approximately 10 km² and required the processing of several thousand street-view images. While the indicators based on object recognition and classification (sidewalk presence, pedestrian entrances and vegetation type) were computed in approximately 2.5 hours on a single A100 GPU, the estimation of street-frontage length required reasoning-intensive visual assessment and took approximately 35 hours. This contrast highlights an important distinction between streetscape indicators that rely primarily on the recognition and counting of visual elements (even if with some spatial reasoning) and those that require extended reasoning, calibration and measurement. These processing times are likely to decrease as the field matures: lower-bit quantization schemes already reduce memory footprint and inference latency with limited accuracy loss, new vision-language model families are released at an accelerating pace with improved efficiency-to-performance ratios, and alternative prompting strategies beyond chain-of-thought (such as structured output grammars, graph reasoning or tool-augmented inference) may reduce the token budget required for measurement tasks without sacrificing accuracy. Nevertheless, the experiment demonstrates that reasoning-based streetscape assessment is already feasible at the scale of a suburban sector and opens the possibility of extending such analyses to entire metropolitan regions covering hundreds or even thousands of square kilometres.

The prospect of scaling up raises the question of operational deployment. Most scoring tasks are inherently parallel and could be distributed across multiple GPUs or computing nodes, making metropolitan-scale coverage a matter of infrastructure rather than methodology. The study suggests that contemporary VLMs can already support the assessment of streetscape qualities directly relevant to urban planning, from pedestrian comfort to vegetation regime and plot accessibility. By enabling the rapid production of fine-grained indicators across extensive suburban territories, SAGAI opens new possibilities for the diagnosis, monitoring and design of pedestrian-friendly environments within the emerging paradigm of the suburban 15-minute city.

## Funding

This work was produced within the emc2 project, funded by ANR (France, grant ANR-23-DUTP-0003), FFG (Austria, grant FO999905461), MUR (Italy, grant 2024/0017648) and Vinnova (Sweden, grant 2023-02581) under the Driving Urban Transition Partnership, which is co-funded by the European Commission.

## Appendix 1

**Shared role**

You analyze a street-level image. Be precise and follow instructions exactly. The street frontage is the street-facing boundary between the public street space and the adjoining plots, on one side of the street only. It may consist of material elements (building facade, fence, wall, hedge) or immaterial elements (edge of a garden, parking lot, vacant plot, or setback area when no physical boundary exists).

**Sidewalk Detection**

*Task:* Detect the presence of a sidewalk delimiting the street frontage visible in the image. *Theory:* The street frontage extends to the first clearly visible perpendicular intersection. If none is visible, stop at the last visible point of the street-facing boundary. Consider as sidewalk both a raised pavement and a protected pedestrian path bordering the street. If the sidewalk does not cover the entire street frontage, still consider it as present.
*Format:* Answer with only one word: yes or no

**Pedestrian Entrance Counting**

*Task:* Count all pedestrian entrances to the plots abutting the street frontage visible in the image. *Theory:* The street frontage extends to the first clearly visible perpendicular intersection. If none is visible, stop at the last visible point of the street-facing boundary. A pedestrian entrance is either: - A door in a building (including shop entrances) giving direct access from the street, or - A pedestrian gate in a fence or wall giving access from the street to an enclosed open area. Garage doors and car only gates are NOT pedestrian entrances. Exception: if a car gate is the only access to a fenced plot and has a doorbell, street number, or mailbox, count it as one pedestrian entrance. If a pedestrian gate gives access to an enclosed open space before a building, count only the gate, not the building doors behind it. Pedestrian entrances partially hidden by foreground objects (trees, cars, people, street furniture) should still be considered if clear visual cues are present.
*Format:* Answer with only one integer number, nothing else.

**Vegetation Classification**

*Task:* Classify the vegetation presence along the street frontage visible in the image into one of six types.
*Theory:* The street frontage extends to the first clearly visible perpendicular intersection. If none is visible, stop at the last visible point of the street-facing boundary. First determine two things: A. Trees in the street space: Trees are in the street space only if located in the street right-of-way along the frontage (on the sidewalk, curb strip, planting strip, or median — between the carriageway and the plot boundary). Trees inside plots (behind the plot boundary) do NOT count as trees in the street space. Only count trees large enough to form a canopy that at least partially covers the street space. If the image was taken in winter, deciduous trees may have no leaves. Still count them as trees if their trunk and branch structure indicate they are large enough to produce a canopy during the growing season. B. Vegetation in plots: Vegetation in plots refers to trees, shrubs, lawn, or hedges seen behind the street-facing boundary inside the abutting plots. In winter, consider deciduous trees and shrubs as vegetation even if leafless. Then select the matching type: Type 1 = Trees in street space YES + Plots highly vegetated (almost entirely covered) Type 2 = Trees in street space NO+ Plots highly vegetated Type 3 = Trees in street space YES + Plots partially vegetated (vegetated setbacks or side gardens) Type 4 = Trees in street space NO + Plots partially vegetated Type 5 = Trees in street space YES + Plots little or no vegetation Type 6 = Trees in street space NO + Plots little or no vegetation.
*Format:* Answer with only one integer number (1 to 6), nothing else.

**Street Frontage Length**

*Task:* Estimate the total length (in meters) of the street frontage visible in the image.
*Theory:* The street frontage extends to the first clearly visible perpendicular intersection. If none is visible, stop at the last visible point of the street-facing boundary. The length is the sum of the linear lengths of all material and immaterial boundary elements along the street, from the left edge to the right edge of the visible street frontage. Only the street-facing boundary line is measured. Elements behind this boundary (such as building facades behind setbacks) are not included in the length, but may be used as visual references. Choose ONE type of standard-size reference object visible near the street-facing boundary and use it consistently to estimate the full length: - Cars (4 to 4.5 m each) - Parking bays (about 5 m each) Doors or windows (about 1 m wide each) - Sidewalk slabs or modules - Building stories projected horizontally (about 3 m per story)- Electricity posts (about 8 m tall) or streetlamps (4 to 8 m tall), projected onto the horizontal boundary line. Account for perspective: objects farther from the camera appear shorter than their actual size.
*Format:* auto-generated by parsing code

## Data Availability Statement

The SAGAI workflow (v2.3, pinning UVLM v4.0.0) is openly available at github.com/perezjoan/SAGAI and permanently archived on Zenodo (https://doi.org/10.5281/zenodo.22028293).